\documentclass[conference]{IEEEtran}
\makeatletter
\def\ps@headings{%
\def\@oddhead{\mbox{}\scriptsize\rightmark \hfil \thepage}%
\def\@evenhead{\scriptsize\thepage \hfil \leftmark\mbox{}}%
\def\@oddfoot{}%
\def\@evenfoot{}}
\makeatother
\usepackage{color, soul}
\usepackage{}
\usepackage{tipa}
\usepackage{bbm}
\usepackage{color}
\usepackage{mathrsfs}
\usepackage{amsfonts}
\usepackage{cite,url,subfigure,epsfig,graphicx}
\usepackage{amssymb,amsmath,bm,makecell}
\usepackage{indentfirst}
\usepackage{float}
\usepackage{algorithm}
\usepackage{algpseudocode}
\usepackage{algorithmicx}

\usepackage{tabulary}
\usepackage{booktabs}
\usepackage{textcomp}
\usepackage{multirow}
\usepackage{tabu}
\IEEEoverridecommandlockouts
\usepackage{times,verbatim,amsfonts,amsmath,color}

\begin{document}

\title{
%Towards Energy-Efficient Edge Intelligence: Federated Learning over 5G+ Mobile Devices
Towards Energy Efficient Federated Learning over 5G+ Mobile Devices
\author{
\IEEEauthorblockN{Dian Shi, Liang Li, Rui Chen, Pavana Prakash, Miao Pan, and Yuguang Fang}
}
\thanks{Dian Shi, Rui Chen, Pavana Prakash, and Miao Pan are with University of Houston; Liang Li is with Xidian University; Yuguang Fang is with University of Florida.}
%\thanks{L. Wang is with the School of Electronic Engineering, Beijing University of Post and Telecommunications, Beijing, China (e-mail: liwang@bupt.edu.cn).}
}

\maketitle\thispagestyle{empty}\maketitle\pagestyle{empty}

\begin{abstract}
The continuous convergence of machine learning algorithms, 5G and beyond (5G+) wireless communications, and artificial intelligence (AI) hardware implementation hastens the birth of federated learning (FL) over 5G+ mobile devices, which pushes AI functions to mobile devices and initiates a new era of on-device AI applications. Despite the remarkable progress made in FL, huge energy consumption is one of the most significant obstacles restricting the development of FL over battery-constrained 5G+ mobile devices. To address this issue, in this paper, we investigate how to develop energy efficient FL over 5G+ mobile devices by making a trade-off between energy consumption for ``working'' (i.e., local computing) and that for ``talking'' (i.e., wireless communications) in order to boost the overall energy efficiency. Specifically, we first examine energy consumption models for graphics processing unit (GPU) computation and wireless transmissions. Then, we overview the state of the art of integrating FL procedure with energy-efficient learning techniques (e.g., gradient sparsification, weight quantization, pruning, etc.). Finally, we present several potential future research directions for FL over 5G+ mobile devices from the perspective of energy efficiency. 

\end{abstract}

\section*{Introduction}

Machine learning (ML), particularly deep learning (DL), is one of the most disruptive technologies the world has witnessed in the last few years. Unfortunately, cloud-centric ML generates tremendous traffic and also causes serious privacy concerns, which is not suitable for many resource-constrained applications. In order to scale and move beyond the cloud-centric ML, Google has introduced federated learning (FL), the currently popular distributed machine learning paradigm, which aims to enable mobile devices to collaboratively learn a joint global ML model without sharing their privacy sensitive raw data \cite{mcmahan2017communication}. With FL, distributed data stakeholders (e.g., mobile devices) only need to periodically upload their updated local models to the aggregation server for global updates, instead of uploading their potentially private raw data, thus significantly lowering the risk of privacy leakage. However, stakeholders in many IoT applications, like smart devices, are resource constrained in terms of spectrum, energy, computing and storage, which makes FL for such on-device applications highly challenging. Therefore, another recent surging technology, 5G and beyond (5G+)~\cite{saad2019vision}, can further facilitate the implementation of FL over mobile devices. First, due to the advance of hardware design, 5G+ mobile devices are usually armed with ever-increasingly high-performance computation units, such as the central processing units (CPUs) and graphics processing units (GPUs), which enable them to host computation-intensive learning tasks. Besides, 5G standard has also embraced the computing capability, such as multi-access edge computing (MEC), paving the way for performing computing for edge intelligence, and hence building an effective wireless network architecture to support viable FL. Moreover, 5G+ wireless transmissions are featured by a very high data rate and ultra-low latency, which can be leveraged to tackle the communication bottleneck issue for local model updates during training. Such a combination of 5G+ and FL prompts tremendous successful applications over 5G mobile devices, including keyboard prediction~\cite{hard2018federated}, cardiac event prediction~\cite{brisimi2018federated}, financial risk management~\cite{yang2019federated}, etc.

While deploying FL over 5G+ mobile devices is promising to have so many interesting applications, FL and 5G+ mobile devices cannot be easily married for fruitful use without frictions. Severe challenges are foreseeable, of which energy consumption is the dominant concern. On the one hand, executing on-device computing and performing local model updates are both resource-hungry, inducing a significant surge of energy consumption on mobile devices and hence draining significant battery power. Thus, the first mountain we have to climb is to improve the system energy efficiency in order to prolong the lifetime of mobile devices during training, where the energy consumption usually comes from both the local computing and wireless communications. On the other hand, for FL over 5G+ mobile devices, there will be a trade-off between computing and communication over resource-constrained mobile devices. This stems from our observation on the comparable energy consumption for on-device training with high-performance processors and wireless transmissions with advanced communication techniques. For example, to transmit a ResNet-50 model, a commonly used deep network for image classification, with approximately 100MB parameters via 100 Mbps 5G wireless uplinks typically consumes 30J for Industry IoT devices~\cite{3gpp.21.915}. This is comparable to the energy consumption for performing a single-step local training on one GPU (e.g., 30J for NVIDIA Tesla V100 on ImageNet dataset~\cite{goyal2018accurate}). Therefore, how to make a trade-off between computing and communications in order to accommodate realistic computing environments is another critical issue when deploying FL over 5G+ mobile devices.

Motivated by the aforementioned challenges, we plan to investigate the energy-efficient FL over 5G+ mobile devices in this paper. Our goal is to enable effective and efficient local training on 5G+ mobile devices while minimizing the overall energy consumption for FL over 5G+ mobile devices for both involved communications and computing. To this end, we first give an overview on the FL over 5G+ mobile devices and discuss the energy consumption models. Then, we study the local computing and wireless transmission co-design from the long-term learning perspective, where we make a trade-off between the two parts simultaneously. In addition, several advanced techniques, including gradient sparsification, gradient quantization, weight quantization, model pruning, and dynamic batch sizes, are integrated into the FL training procedure to further reduce energy for 5G+ mobile devices. Finally, we conclude this paper with discussions on potential research directions for energy efficient FL over 5G+ mobile devices.

\section*{Backgrounds and Energy Models}

\subsection*{FL over 5G+ Mobile Devices}
\begin{figure}[t]
\centering
\includegraphics[scale=0.29]{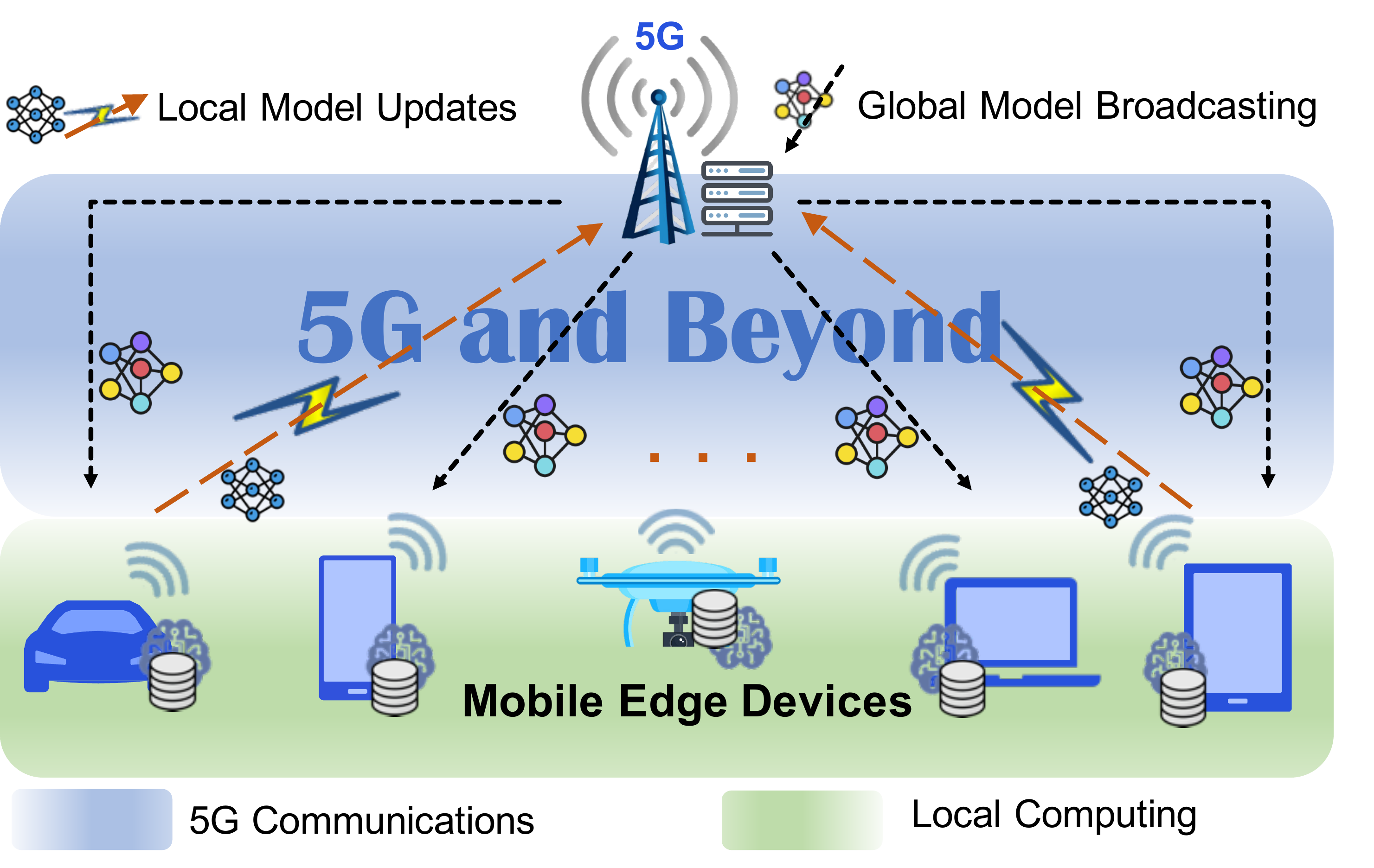}
\caption{The illustration of the FL over 5G+ mobile devices.}
\label{fl_frame}
\end{figure}

As an emerging decentralized learning paradigm, FL has taken advantage of the computing resources across massive participants. Specifically, all participants collaboratively contribute to one global learning task in a distributed manner with continuous interactions for model parameter updates. With FL over 5G+ mobile devices, all 5G+ mobile devices (e.g., smartphones, laptops, automatic vehicles, etc.) can serve as participants, and a server such as gNodeB acts as the aggregator. In particular, a 5G server first broadcasts a current global model to the participating 5G+ mobile devices in FL. After receiving the global model, a 5G+ mobile device conducts the local on-device training based on the local data and its computing capability. Second, when a 5G+ mobile device finishes its local training in this round, it will upload its local model updates (i.e., gradients) to the server via wireless links with 5G+ techniques. Finally, the server does the aggregation over all the received local gradients to update the global model and then feeds it back to the participated mobile devices for the next-round training. The above procedures are repeated until obtaining a converged global model, which can be deployed by the 5G+ devices for future utilization. A typical paradigm of the FL over 5G+ mobile devices, including local computation and wireless communications parts, is shown in Fig.~\ref{fl_frame}.

FL over 5G+ mobile devices has become a natural way to implement the artificial intelligence (AI) at the edge, like the keyboard prediction introduced by Google~\cite{hard2018federated}. Such a combination pushes AI functions to mobile devices, which provides a flexible and convenient approach to conducting a learning task, especially for some real-time and lifelong learning applications. However, deploying FL over 5G+ mobile devices raises tremendous challenges and difficulties, and energy consumption is one of the most significant issues. Unlike other central servers with wired connections, mobile devices have limited energy resources due to the limited battery power. Besides, the energy consumption of wireless transmissions is not encountered in learning scenarios with wired servers. Both make it extremely difficult for 5G+ mobile devices to handle the energy-hungry training tasks. In light of this, it is worthwhile to investigate the FL's energy model over 5G+ mobile devices to deal with energy-saving issue.

\subsection*{Communication and Computation Energy Models}

With FL, all 5G+ mobile devices contribute to one unified global model by continuously transferring the local model updates with the centralized aggregator. In this process, the energy consumption of the 5G+ mobile devices mainly comes from the wireless transmissions of the model updates and the local computation executed on them. Hence, brief descriptions of the communication energy model and the computation energy model are given as follows.

\textbf{Communication:}
All the participating 5G+ mobile devices transmit their computed local model updates to the central aggregator through the wireless transmissions, which corresponds to the communication energy consumption. Note that conducting an FL task usually takes several minutes due to the heavy computational load of training. In this situation, the channel condition may not remain the same all the time and may suffer from fluctuations. Therefore, one possible way to model the transmission rate for each device throughout the entire training process is to do it in an expected manner, where the expectation is taken over the channel variations. Besides, according to the Shannon–Hartley theorem, both the bandwidth and the transmission power also impact the transmission rate, and the power of additive white Gaussian noise (AWGN) needs to be considered as well. Accordingly, the overall energy consumed by each device for wireless transmissions during the training process can be formulated as the product of the required number of global communication rounds and the energy consumption in a single round. Here, the one-round energy consumption is related to the transmission power, the transmission rate, and the model size.

According to the characteristics of the energy model mentioned above, two possible ways can be adopted to save the energy consumption on the communication part in the entire training process, namely, decreasing the required rounds of global communications and reducing the communication workloads per round. Given this observation, several approaches can be implemented to save the communication overhead. On the one hand, global synchronizations can be taken after several local computing iterations to decrease the communication frequency, that is, the federated average~\cite{mcmahan2017communication}. Similarly, the number of required global communication rounds can also be reduced by gradually increasing the batch size throughout the training. On the other hand, model compression technologies, e.g., model sparsification and quantization, can greatly help reduce the size of the local model to be transmitted, thus saving the communication energy in each round.

\textbf{Computing:}
With the ever-increasing popularity of smart devices equipped with high-performance GPUs, 5G+ mobile devices can undertake heavy computations even for deep learning tasks. However, due to the powerful computational capability and the massively parallel architecture of the GPU, the computational energy consumption has been a significant burden for the learning scenarios, especially for training tasks implemented on the mobile devices with limited battery power. Thus, more research efforts are needed to investigate the computing energy model and the corresponding computing energy-saving strategies. Here, we assume that the 5G+ mobile devices are equipped with GPUs, which are widely assumed in modern learning training. Specifically, the GPU computation architecture involves the memory modules referring to data fetching and the core modules referring to the data calculation. Under this architecture, the voltage and the frequency of the corresponding modules can be controlled independently. 

The energy consumption for feeding a mini-batch of data in one local iteration can be calculated as the product of the execution time and the runtime power. Here, the execution time is determined by the device-dependent parameters, like the memory frequency and the core frequency, and the task specific information, such as the number of cycles for data fetching and calculation~\cite{mei2017energy}. Similarly, the runtime power is also affected by device-dependent parameters, including the frequency and the voltage, and coefficients related to the specific learning tasks. Hence, the total computation energy consumption can be computed as the product of energy consumption for one local iteration and the total number of iterations. Currently, model compression techniques, such as pruning and weight quantization, together with the corresponding hardware co-design, can  alleviate the burden of computing energy consumption in one local iteration because the needed number of cycles for data fetching and calculation is decreased. The total number of local iterations will increase mildly due to the falling of the model precision in each local iteration, but in all, the total computational energy will be saved.

\section*{Energy-Efficient FL via Communication and Computing Co-Design}

\begin{figure}[t]
\centering
\includegraphics[scale=0.25]{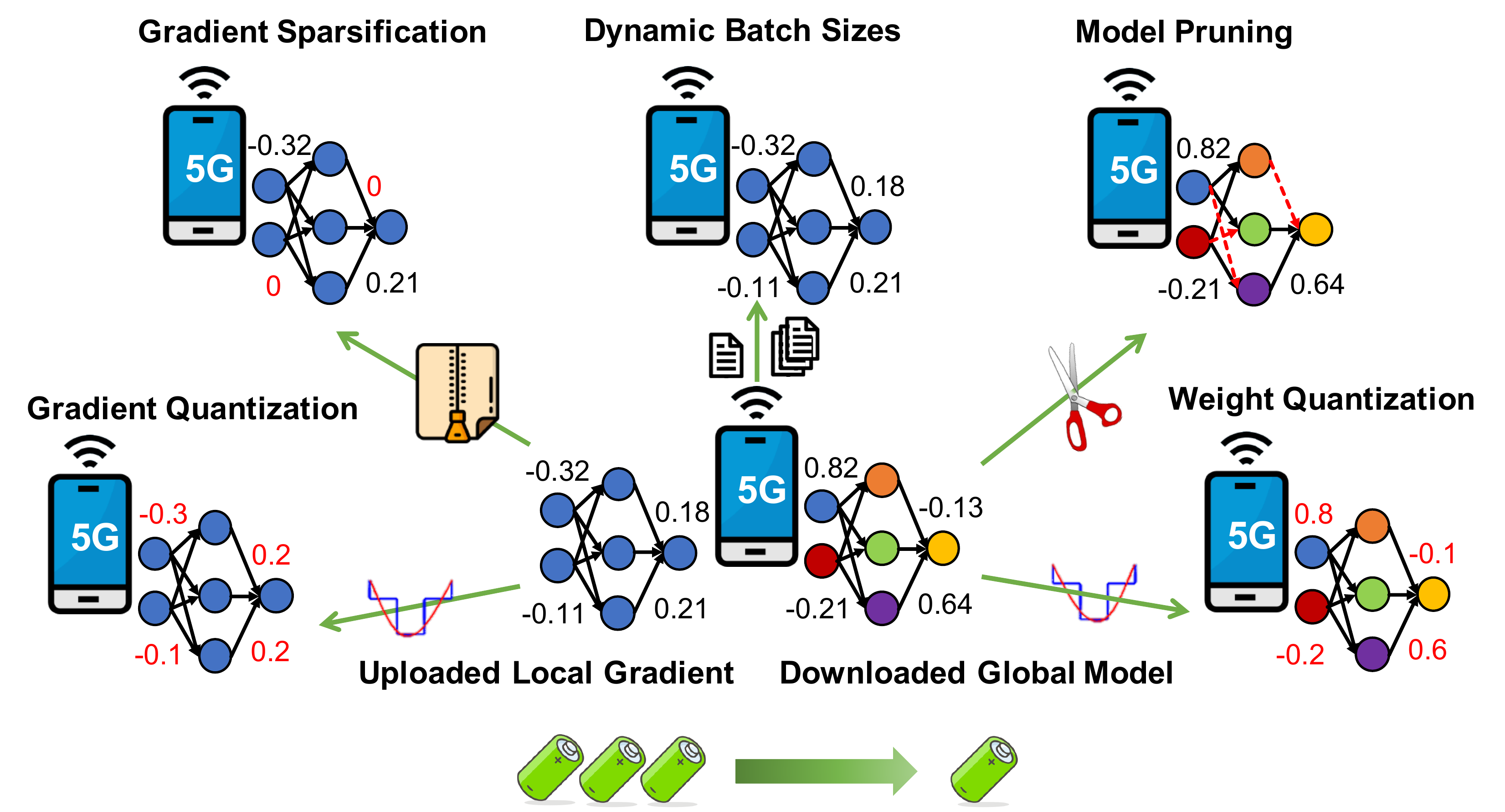}
\caption{Illustrative scenarios for energy-efficient FL over 5G+ mobile devices.}
\label{Fig:energy_eff}
\end{figure}

As illustrated in the previous section, the energy consumption of FL over 5G+ mobile devices mainly comes from two parts: on-device local computing and wireless communications for model updates. Thus, in this section, we focus on how to integrate various technologies (e.g., gradient sparsification, gradient quantization, weight quantization, model pruning, etc.) and develop corresponding communication and computing co-design to reduce overall energy consumption for FL over 5G+ mobile devices. In this section, we will survey some recent developments focusing on the communication and computing energy consumption co-design, which are illustrated in Fig.~\ref{Fig:energy_eff}.

\subsection*{Gradient Sparsification to Reduce Communication Energy}

FL communication energy consumption can be reduced by integrating the training algorithm with two state-of-the-art communication compression strategies, namely, local computations and gradient sparsification. The former allows more local computations performed on the 5G+ mobile device between every two global model updates, thereby reducing the total number of communication rounds. The latter lets participants upload only a fraction of gradients with significant magnitudes, thereby reducing the communication payload in each round. Besides, error compensation is applied at each participant after every model update to accelerate the global convergence by accumulating the error that arises from only uploading sparse approximations of the gradient updates, ensuring all gradient information gets eventually aggregated.

\begin{figure} \centering %\hspace{-3em}
  \subfigure[Test error vs. energy consumption.\label{Fig:MC_enr}]
  {\includegraphics[width=.35\textwidth]{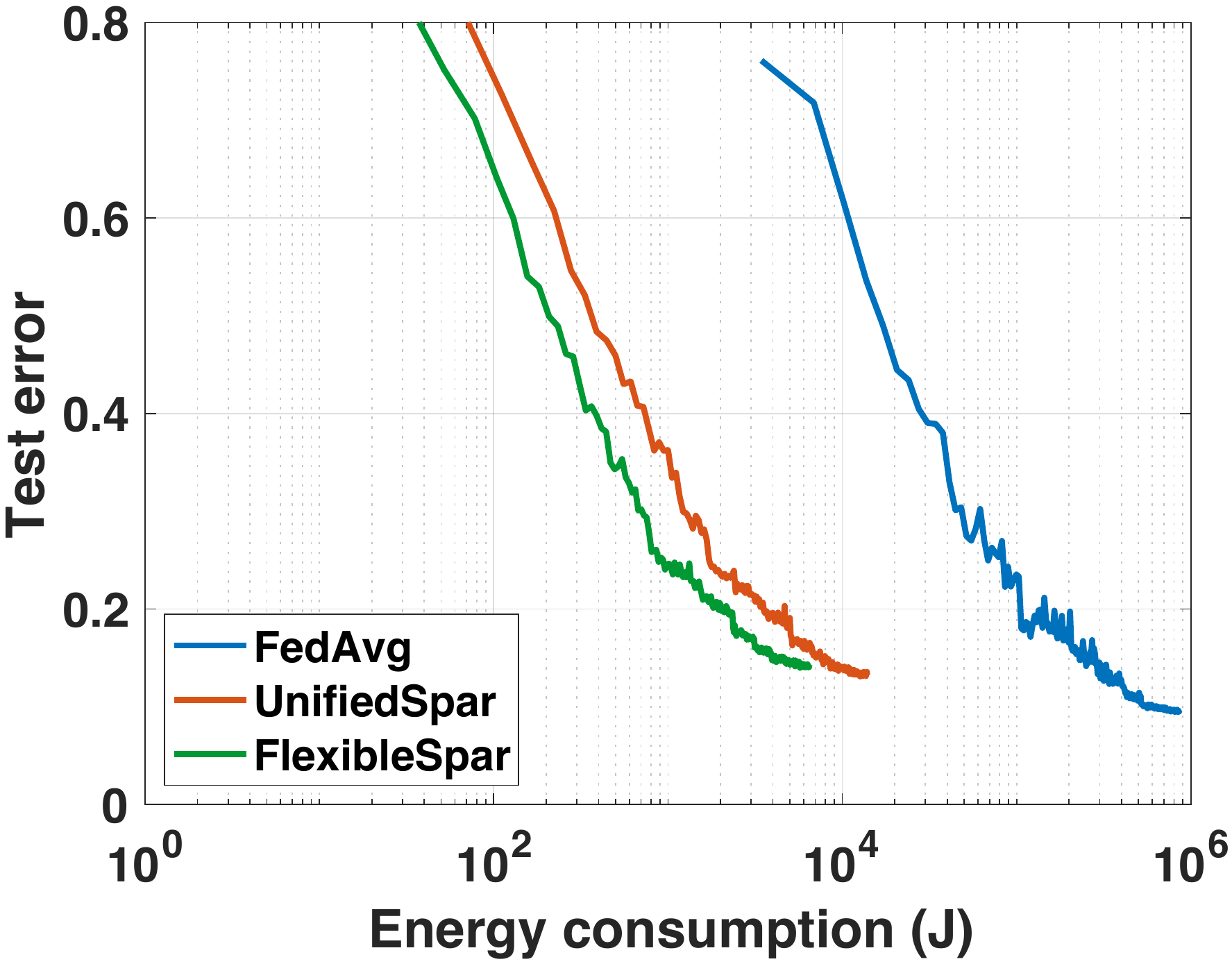}}
  \subfigure[Training accuracy vs. epoch.\label{Fig:MC_epo}]
  {\includegraphics[width=.35\textwidth]{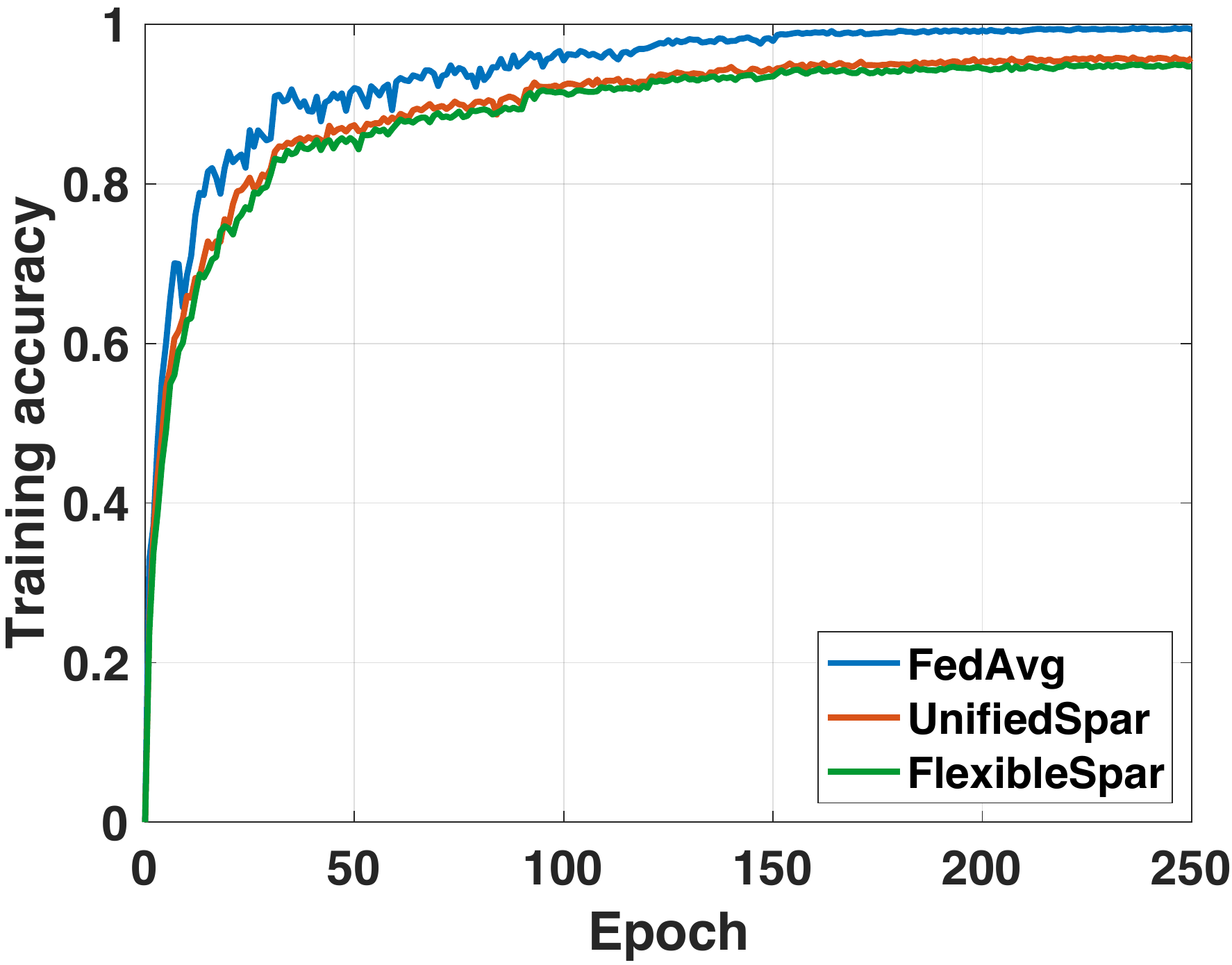}}
  \caption{Performance evaluation for the FL with gradient sparsification.} \label{Fig.MC}
\end{figure}

The convergence results for our FL algorithm indicate that, from the learning perspective, the gradient sparsity magnitudes of all the participants jointly make an impact on global convergence and communication complexity. Given a target model accuracy, a lower sparsity results in a larger bound of communication rounds. Besides, aggressively increasing local iterations can also impair the learning efficiency since more communications may be involved. In a realistic edge computing environment, these two types of compression factors implicitly determine the energy consumption for participating 5G+ mobile devices by affecting the payload required for transmissions and the workload required for processing, respectively. Thus, they are needed to be well balanced for each participant to accommodate the specific environment. This can be achieved by formulating a compression control problem using the derived convergence bound from the long-term learning perspective~~\cite{Li2021infocom}, with the goal for optimizing the overall energy efficiency for FL on 5G+ mobile devices over wireless networks. As shown in Fig.~\ref{Fig:MC_enr}, the flexible sparsification based FL method (``\textit{FlexibleSpar}''), which considers the heterogeneity of participating 5G+ mobile devices and provides the flexible sparsification strategies, consume less energy than the other methods (FedAvg; ``\textit{UnifiedSpar}'': the unified sparsity approach) to reach a given target accuracy. Furthermore, Fig.~\ref{Fig:MC_epo} shows the convergence rate in terms of training epochs, which indicates that ``\textit{FlexibleSpar}” exhibits very similar behavior with ``\textit{UnifiedSpar}” in terms of convergence rate and final accuracy, both of which slightly underperform the baseline approach, i.e., FedAvg~\cite{mcmahan2017communication}.

\subsection*{Gradient Quantization to Reduce Communication Energy}
Similar to the gradient sparsification technique, another efficient compression method to reduce the communication energy consumption is gradient quantization. After completing the local training in one round, the computed gradients should be uploaded to the aggregator for the global updates. With gradient quantization, instead of uploading all raw gradients, we can also quantify the computed local gradients with low precision, i.e., a small number of bits, thus reducing the communication load per round. In this way, a high gradient quantization level corresponds to a low communication energy consumption per round. In contrast, the required global rounds increase due to the precision reduction, thus increasing the total computing energy. Therefore, the gradient quantization level needs to be controlled to make a trade-off between the communication and computing for the overall energy efficiency. It should be noted that local updates (i.e., gradients) refer to the communication part, and thus gradient quantization can reduce the communication energy consumption. However, when we download the global model on the device, the weight quantization is related to the computing part, which will be discussed in the following subsection.

\subsection*{Weight Quantization to Reduce Computing Energy}

Weight quantization is regarded as a promising solution to decrease the local memory consumption in FL training. It reduces the model complexity and computing energy consumption via representing the model parameters with small bit-widths (e.g., 8-bit or 16-bit fixed point numbers) on 5G+ mobile devices. Besides, considering device heterogeneity, the quantization selections for different participants are varying. However, the quantization induces information loss during training. The errors between the quantized and original values make the FL model converge to a neighborhood of the optimal solution. Smaller quantization levels (i.e., bit representations) lead to a higher error and push the FL model further away from the minima. For synchronous aggregation, given a fixed number of communication rounds, the accuracy of the FL model depends on the average quantization errors of all the participants. In this way, given a model accuracy, the participants, with limited computing power and poor channel conditions, prefer more aggressive quantization strategies (smaller bit-widths) to train the local models efficiently. In contrast, others would coordinate with higher precision training for the model performance.

\begin{figure} \centering %\hspace{-3em}
  \subfigure[Training accuracy vs. energy consumption.\label{Fig:NQ_enr}]
  {\includegraphics[width=.35\textwidth]{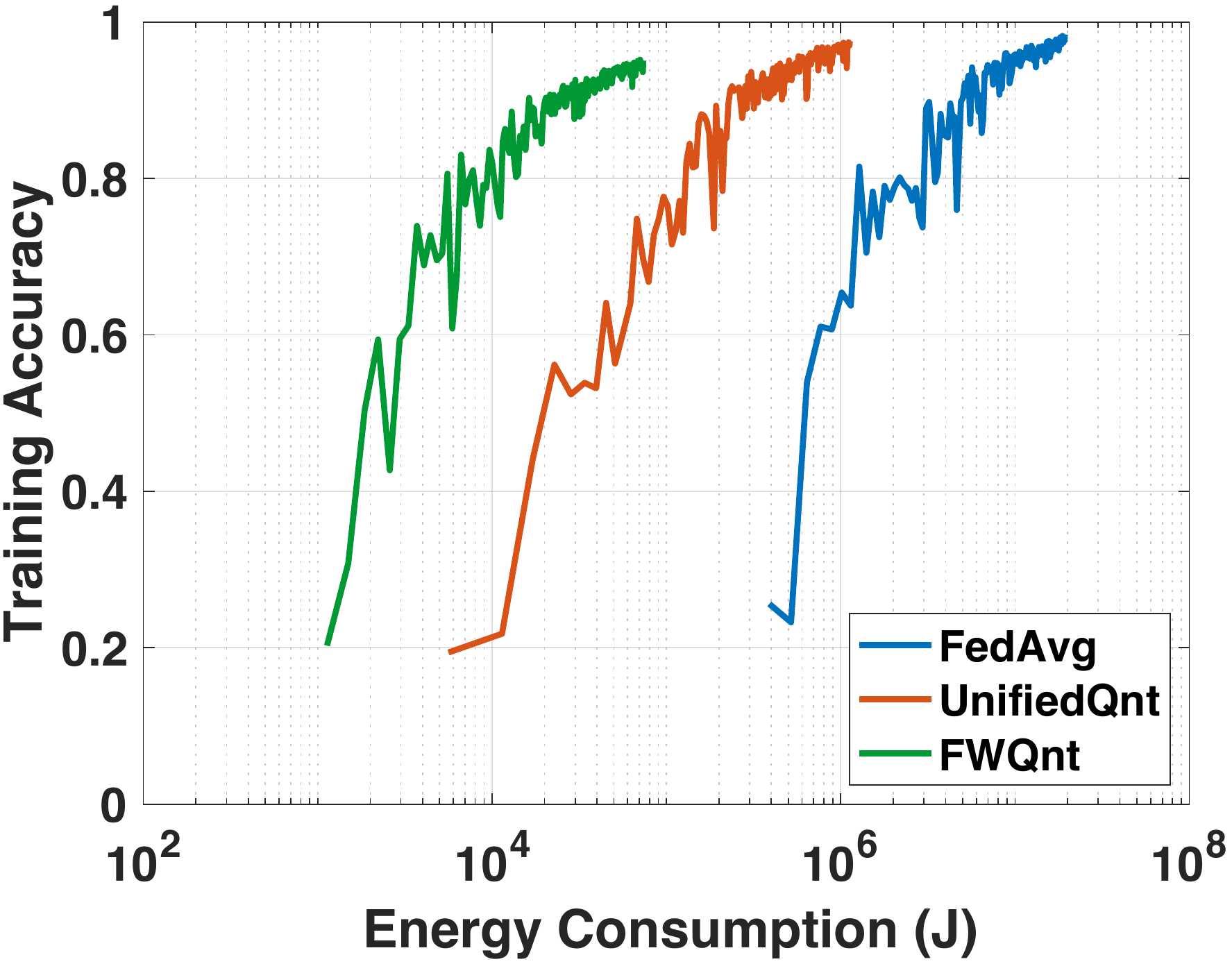}}
  \subfigure[Test error vs. epoch.\label{Fig:NQ_epo}]
  {\includegraphics[width=.35\textwidth]{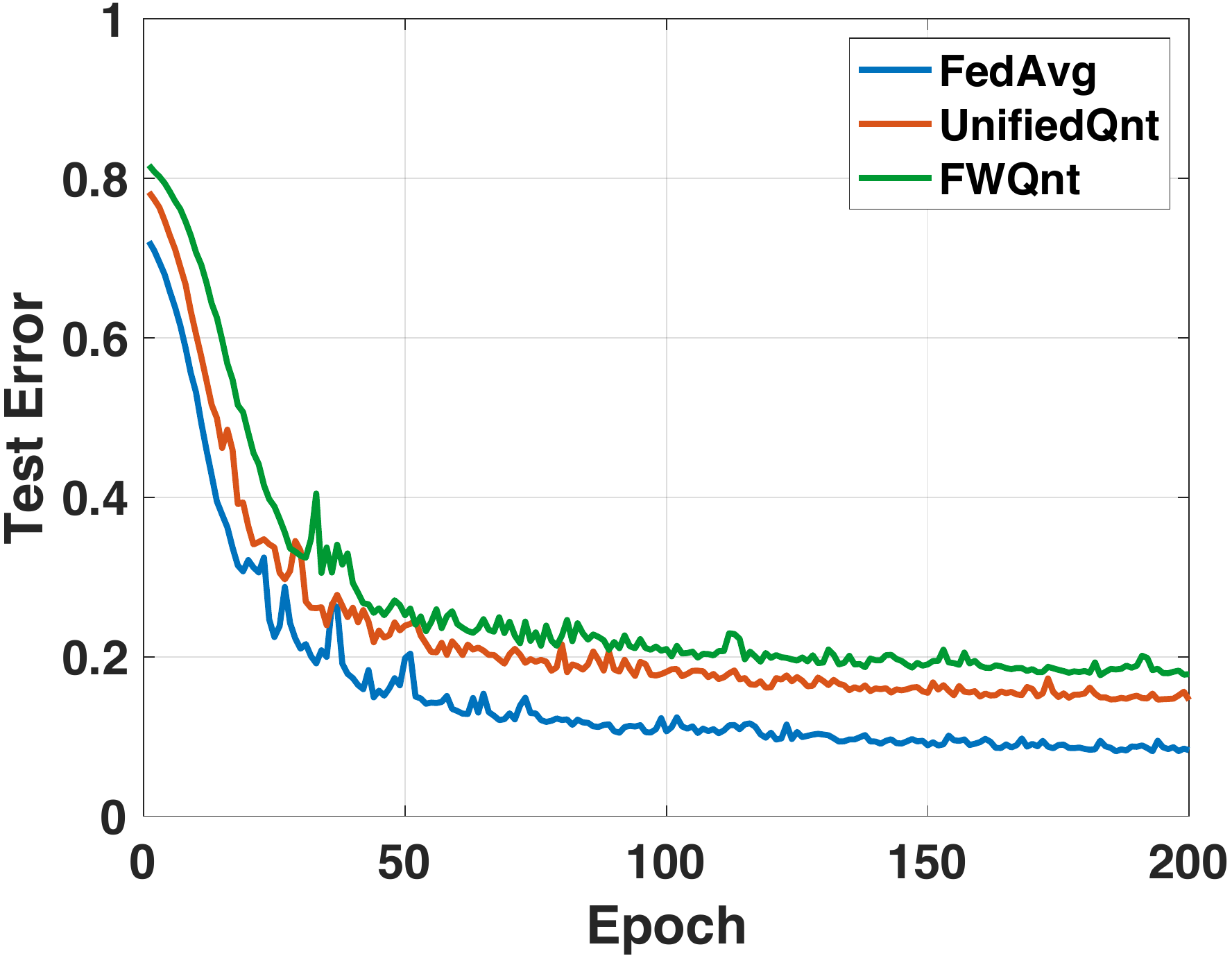}}
  \caption{Performance evaluation for the FL with weight quantization.} \label{Fig.NQ}
\end{figure}

As a result, one can consider a joint design for flexible quantization selection and bandwidth allocation to capture the trade-off between the local computing and communications and minimize the overall energy consumption for FL training within allowed deadline~\cite{Chen2021infocom}. One example of a stochastic quantization scheme could first determine a quantization set based on different quantization levels, and then map the model weights to the nearest quantization point with high probability. The participating devices could determine different quantization levels depending on their device capabilities and the targeted model accuracy. Meanwhile, the server allocates the wireless bandwidth to the participants considering both the channel conditions and participants’ computing capabilities. This process terminates when it reaches a certain global model accuracy level. Fig.\ref{Fig:NQ_enr} shows the overall energy consumption for the FL training procedure under different learning mechanisms. For a fixed number of training iterations, those mechanisms equipped with quantization (“UnifiedQnt” and “FWQnt”) consume less energy than ``FedAvg" without quantization. Specifically, the scheme “FWQnt” considers the model performance, device heterogeneity and wireless channel conditions and enhances the energy efficiency with x2 - x100 less energy consumption than the other schemes in the FL training process under the same accuracy level. Moreover, the convergence rates for the corresponding schemes are also shown in Fig.\ref{Fig:NQ_epo}, where both quantization based approaches slightly underperform ``FedAvg".

\subsection*{Model Pruning to Reduce Computing Energy}

Another efficient learning technique that can be integrated into the FL process is model pruning, which can save the computing energy during training with proper underlying code design. Generally speaking, model pruning compresses the model by removing less contributing weights and connections, while retaining the performance of the original dense models. This will significantly affect the reduction of the computing energy per iteration required. When utilizing gradual pruning schemes, with the model size gradually reduced during training, energy consumption due to communications is also decreased. Besides, reducing the model size also saves the inference time and energy, enabling the pruning method to be much more suitable for FL training over 5G+ mobile devices.

Moreover, the pruning can be coupled with the quantization technique in FL training to address the restrictions in implementing deep neural networks on resource-constrained 5G+ mobile devices. Essentially, quantization requires a smaller number of bits to represent each pruned connection, thereby reducing memory, bandwidth and energy consumption. To enable a communication efficient and mobile device compatible FL process, we develop a three-fold compression of double quantization along with model pruning approach~\cite{prakash2021infocom}. In particular, the gradients and weights of uplink and downlink models are appropriately quantified, and the gradual pruning of the received model is utilized to reduce the computing and communication loads. This has a dual advantage of reducing the communication time and energy, and reducing the memory bandwidth due to less memory accesses. Therefore, we can reduce the model redundancy and make the FL process computation, storage, and communication sufficiently efficient to deploy large-sized deep neural networks over 5G+ mobile devices.

\subsection*{Other Techniques to Save Energy}

The total energy consumption can also be reduced by capturing the intrinsic training dynamics, such as dynamically adjusting the batch size~\cite{shi2021infocom}. Specifically, we can interpret the stochastic gradient descent (SGD) training process as integrating a stochastic differential equation (SDE) whose ``noise scale" is related to the batch size selection. Small batch size theoretically corresponds to large-scale random fluctuations, which can help explore the parameter space to avoid trapping in local minima at the initial stage in the FL problem. At later stage, small-scale fluctuations (large batch size) are more desirable to fine-tune the parameters when a promising region of parameter space is reached. Therefore, gradually increasing batch size in the training process with a well-designed increment strategy can help reduce the communication rounds in FL training process. 

Moreover, gradually increasing batch size also leads to positive effects on computational energy-saving. Due to the GPU's parallelism property, the local computing energy is no longer proportional to the batch size. Accordingly, the energy consumption of unit data calculation is relatively small for large batch training, especially executing on the 5G+ mobile devices with multiple GPUs. Besides, it has been theoretically and experimentally demonstrated that both the fixed batch size approach and the dynamic batch size approach need similar data epochs. In this case, thanks to the benefit of large batch training at later stage, the training approach with dynamical batch sizes is more energy-efficient. Moreover, a batch size control scheme catering to the GPU computing performances and wireless communication conditions of mobile devices can be further developed to balance the computing and communications, thus achieving energy-efficient FL over 5G+ mobile devices.

\subsection*{Balancing Communications and Computing to Reduce Overall Energy Consumption}

As aforementioned, the trade-off always exists between communications and computing. Therefore, it is widely expected that both computing and communication parts need to be considered in the energy-efficient FL training process. When integrating both in the FL training process, the needed number of global communication rounds is a critical element. One possible way to approximate the needed global rounds is to conduct the FL theoretical convergence analysis. After identifying the specific dataset and the training model, the bound of the required number of communications can be derived based on the required training accuracy under some mathematical assumptions. Note that some learning settings, e.g., batch size, local iteration numbers, etc., also impact the required number of communications. Accordingly, the total energy consumption corresponding to the required number of global rounds can be obtained, which exhibits a global view of the overall energy consumption in the FL training process. Despite employing the computing or communication efficient learning techniques, the energy efficiency cannot be significantly improved without elaborately controlling the key training parameters from a global perspective. This requires us to optimize the training parameters, so that communications and computing can be well balanced. This is why, in the above subsections, we firstly study the efficient communication (e.g., gradient sparsification, gradient quantization and dynamic batch sizes) or efficient local computing (e.g., weight quantization and pruning~\cite{han2015deep}) methods in order to find the effective schemes. We then integrate these efficient methods with resource allocation strategies to strike a good balance between communications and computing, thus minimizing the energy consumption for FL over 5G+ mobile devices.

\section*{Challenges and Future Research Directions}

Although there are a few pioneering research works done on the energy efficiency for FL over 5G+ mobile devices, the relevant study is still in its infancy and requires more thorough investigation. In this section, we summarize some existing challenges and potential future directions.

\subsection*{Energy-Efficient FL over Heterogeneous Mobile Devices}

Most of the existing energy-efficient FL algorithms assume certain homogeneity of mobile devices, while they may be considerably different in practice. The heterogeneity across participating devices may lie in local training data, computing capability, and wireless channel condition, etc., which require flexible and customizable training strategies for each participant. Thus, it is expected to develop advanced design methodology for efficient FL over 5G+ mobile devices across multi-dimensional heterogeneity. Accordingly, the quantization granularity, the pruning strategy, and compression levels for different participants can vary to accommodate their realistic environments for energy-saving.  For example, the device with powerful GPUs but poor channel conditions may choose to compute more and communicate less and vice versa. In these situations, it is also significant to optimize the personalized compression or pruning strategy, adapting to the heterogeneous devices' capability for total energy saving by balancing communications and computing.

\subsection*{Energy-Efficient FL under Flexible Aggregation}

The practical scenarios with heterogeneous data and devices demand higher requirements on the aggregation strategies. On the one hand, the updated local models across devices may differ in size or even structure, which invalidates the current FL aggregation schemes (e.g., FedAvg~\cite{mcmahan2017communication}) if averaging model parameters directly. Thus, we need to investigate more powerful and more flexible aggregation schemes for FL over heterogeneous 5G+ mobile devices, while mitigating the energy consumption to the same extent as the baseline FL algorithms. On the other hand, due to the Non-IID (Independent and Identically Distributed)  data sources and different precision for local models, each participant may have different contributions to the global model in terms of accuracy and convergence. Therefore, from the energy perspective, only a proportion of the participants need to contribute their models in each communication round, or the aggregation will be performed in an asynchronous way, thus reducing more energy compared with the original inefficient aggregation method. Overall, flexible and asynchronous aggregations are efficient methods to further reduce energy consumption, which deserves further investigation.

\subsection*{Energy-Efficient FL with Privacy Preservation}

One of the inherited features of FL over 5G+ mobile devices is the privacy preservation of the users' sensitive raw data. Unfortunately, the private information can still be inferred from local updates communicated between user devices and the aggregation server with some recently developed attack mechanisms. Therefore, such a distributed learning, which needs to exchange intermediate model parameters, brings in a significant design challenge for privacy protection. A common strategy to enhance the privacy and security for participating users is to introduce perturbations into the FL training framework, such as adding the noise. Fortunately, some early research works have already shown that pruning, quantization, and other energy-efficient methods can introduce randomness into the FL training process and provide the enhanced privacy guarantee, i.e., intrinsic privacy preservation. Under such conditions, energy-saving strategy and privacy protections are perfectly combined, where the privacy will be preserved without additional noisy computation. However, the related research is still in its initial stages and needs deeper exploration, especially for the differential privacy preservation, where the privacy level can be precisely quantified.

\subsection*{Extensive Applications of Energy-Efficient FL}

With the massive growth of personal data with end users and the rapid popularization of the power-efficient mobile edge devices, FL over 5G+ mobile devices can be applied to a large number of applications~\cite{li2020review} and can be involved in every area of daily life. With energy-efficient training strategies, FL over 5G+ mobile devices is perfectly compatible with lifelong on-device learning that requires a constant training process and is battery-driven. For example, it can be applied to some real-time assisting services like the voice UI, keyboard prediction, and some low-latency control scenarios such as gaming and automated guided vehicles~\cite{Fang2019ieee}. Moreover, along with the exponential improvement in on-device AI capabilities, more sensing data from smart sensors like the cameras, microphones, and compass, can be effectively utilized in Industrial IoT, e-health, finance, and social networks, etc.

\section*{Conclusion}

This paper has studied the FL over 5G+ mobile devices to address the issue on energy consumption during the FL training process. We have investigated how to properly conserve energy and allocate resources during FL training. We start with introduction of wireless communications and on-GPU computing models. Then, we discuss several energy-efficient training techniques, including gradient sparsification, gradient quantization, weight quantization, model pruning, and dynamic batch sizes, to save energy. At the same time, the resource allocation strategies are adapted to reasonably manage energy resources by balancing communications and computing energy consumption. We conduct extensive simulations to demonstrate the efficacy of the techniques mentioned above for FL over 5G+ mobile devices. Finally, we have presented some existing design challenges and the corresponding research directions.


%\bibliographystyle{IEEEtran}
%\bibliography{main.bib}

\end{document}